\documentclass[letterpaper,10pt,conference]{ieeeconf}

\IEEEoverridecommandlockouts
\usepackage{amsmath,amssymb}
\usepackage{balance}
\usepackage{booktabs}
\usepackage{graphicx}
\usepackage{microtype}
\usepackage{multirow}
\usepackage{stfloats}
\usepackage{tikz}
\usepackage{xcolor}
\usepackage{xspace}
\usepackage{url}
\usetikzlibrary{arrows.meta,fit,patterns,positioning}

\definecolor{aquablue}{RGB}{20,112,145}
\definecolor{aquagreen}{RGB}{28,137,117}
\definecolor{softgray}{RGB}{225,229,232}
\newcommand{\method}{AquaJEPA\xspace}
\newcommand{\methodbase}{AquaJEPA-base\xspace}
\newcommand{\methodrobust}{AquaJEPA-robust\xspace}
\newcommand{\methodcamera}{AquaJEPA-C\xspace}
\newcommand{\methodsonar}{AquaJEPA-S\xspace}

\newcommand{\TotalClosedLoopRuns}{720}
\newcommand{\SeedwiseClosedLoopRuns}{2160}

\newcommand{\ClosedLoopAbstractSentence}{\methodbase achieves the strongest
aggregate closed-loop performance, improving success over state-only by 12.5
percentage points and reducing final error by 0.189~m; both paired 95\%
intervals exclude zero. In a separate three-seed evaluation, it reduces paired
final error relative to \methodsonar by 0.118~m, with the same direction for
every seed. \methodrobust more than halves prediction error during camera and
camera--DVL blackouts. These results show that full multimodal prediction
improves over state-only control and the sonar-only family member in this
benchmark, while sensor-dropout training provides robustness under sensor
loss.}

\newcommand{\PredictionResultSentence}{\methodbase and \methodrobust reach
0.647 and 0.634 action rank on the 15-episode held-out split. Removing the
action margin from the robust configuration lowers rank to 0.573 despite
slightly reducing velocity MAE from 0.0032 to 0.0030, supporting the margin's
role in counterfactual action discrimination. Supervised dynamics attains the
lowest physical prediction errors, showing that short-horizon MAE alone does
not determine the closed-loop ordering.}

\newcommand{\CorruptionResultSentence}{Under camera blackout,
\methodrobust retains 0.0032 velocity MAE, whereas its matched
no-modality-dropout ablation rises to 0.0088; the combined camera--DVL
condition gives 0.0037 versus 0.0090. Thus training with missing modalities
cuts error by more than half in both interventions. The no-explicit-mask
ablation matches the robust configuration on this fixed-corruption metric, so
the test does not isolate an additional mask benefit.}

\newcommand{\ClosedLoopResultSentence}{\methodbase records the strongest
aggregate result at 91/120 successes and 0.617~m final error. Against
state-only (76/120 and 0.806~m), its paired success difference is $+0.125$
(95\% CI $[+0.075,+0.183]$) and its paired final-error difference is
$-0.189$~m ($[-0.239,-0.140]$). \methodrobust reaches 84/120 successes and
0.766~m final error, trading pooled closed-loop performance for the explicit
blackout robustness isolated in Table~\ref{tab:corruption-ablation}.}

\newcommand{\StratifiedResultParagraph}{The success advantage over state-only
appears at every visibility coefficient: \methodbase reaches 23, 24, 22, and
22 of 30 goals from Jerlov 0.10 through 0.25, versus 20, 20, 17, and 19. Under
nominal dynamics it reaches 60/60 goals versus 58/60 for state-only; under the
shift it retains 31/60 versus 18/60. By layout, \methodbase reaches 39/40 goals
in east corridor, 23/40 in north slalom, and 29/40 in open arc, compared with
32/40, 23/40, and 21/40 for state-only.}

\newcommand{\ConclusionResultSentence}{Across 120 fresh paired
poor-visibility environments, \methodbase achieves the best aggregate result,
improving closed-loop success over state-only by 12.5 percentage points (95\%
CI: 7.5--18.3) and reducing final error by 0.189~m (95\% CI: 0.140--0.239),
while \methodrobust more than halves prediction error during camera and
camera--DVL blackouts. Across three independently trained checkpoints,
\methodbase also reduces final error relative to \methodsonar by 0.118~m
(hierarchical paired 95\% CI: 0.047--0.181), consistently across seeds.}

\title{\LARGE \bf AquaJEPA: An Action-Conditioned Multimodal JEPA Family\\
for Underwater Robot Dynamics}

\author{Alan-Barsag Gazzaev, Alexey Gavrilov, and Sergey Muravyov%
\thanks{All authors are with ITMO University.}%
\thanks{This work has been submitted to the IEEE for possible publication.
Copyright may be transferred without notice, after which this version may
no longer be accessible.}}

\begin{document}
\maketitle
\thispagestyle{empty}
\pagestyle{empty}

\begin{abstract}
Underwater robots rely on complementary sensors whose reliability changes
abruptly with water visibility and vehicle motion. We introduce \method, a
sensor-configurable family of action-conditioned joint-embedding predictive
models spanning full multimodal, camera-only, sonar-only, and sensor-dropout
configurations. Its members share a latent objective and receding-horizon
control interface that predict future representations and physical dynamics
from camera, forward-looking sonar, proprioception, and thruster commands.
Trained from scratch on one hour of action-labelled data, the family is
evaluated in Stonefish on 120 fresh paired scenarios spanning unseen layouts,
visibility changes, dynamics shifts, and scheduled DVL loss.
\ClosedLoopAbstractSentence{}
\end{abstract}

\section{Introduction}

Underwater perception is intrinsically heterogeneous. An RGB camera provides
texture and semantics but degrades with attenuation, backscatter, and
illumination. Forward-looking sonar (FLS) retains geometric returns in
conditions where optical sensing is weak, yet offers lower angular and semantic
resolution. IMU, pressure, and Doppler velocity log (DVL) measurements add
motion information, but DVL bottom lock can disappear. A dynamics model that
assumes a complete observation vector can therefore be least reliable in the
conditions where model-based control is most valuable.

World models compress observations and predict how actions change future
states \cite{ha2018world,hafner2019planet,hafner2020dreamer}. Pixel-level
prediction, however, allocates capacity to appearance details that may be
unpredictable and irrelevant to control. Joint-embedding predictive
architectures (JEPAs) instead predict future features
\cite{lecun2022path,assran2023ijepa,bardes2024vjepa}. For underwater control,
the representation must additionally distinguish futures caused by different
thruster sequences and remain interpretable when one sensing stream vanishes.

\begin{figure*}[t]
\centering
\begin{tikzpicture}[
    font=\scriptsize,
    node distance=4mm and 4mm,
    box/.style={draw, rounded corners=1.5pt, minimum height=6.5mm, align=center,
                fill=white, inner xsep=3pt},
    sensor/.style={box, fill=softgray},
    learned/.style={box, draw=aquablue, very thick, fill=aquablue!7},
    target/.style={box, draw=aquagreen, thick, dashed, fill=aquagreen!7},
    arrow/.style={-{Latex[length=2mm]}, thick},
    loss/.style={-{Latex[length=2mm]}, dashed, aquagreen, thick}
]
\node[learned] (ecam) {camera encoder\\$E_I$};
\node[learned, right=2mm of ecam] (esonar) {sonar encoder\\$E_S$};
\node[learned, right=2mm of esonar] (eprop) {proprio encoder\\$E_p$};
\node[sensor, above=of ecam] (cam) {RGB\\$I_t$};
\node[sensor, above=of esonar] (sonar) {FLS\\$S_t$};
\node[sensor, above=of eprop] (prop) {IMU, DVL, depth\\$p_t$};

\node[draw, rounded corners=2pt,
      fit=(cam)(sonar)(prop)(ecam)(esonar)(eprop), inner sep=4pt,
      label={[font=\scriptsize]below:multimodal observation encoder}] (obs) {};

\node[learned, right=7mm of obs] (fusion)
    {mask-aware fusion\\$F(h_t^I,h_t^S,h_t^p,m_t)$};
\node[sensor, above=of fusion] (mask) {validity mask\\$m_t$};
\node[learned, right=7mm of fusion] (pred)
    {joint predictor\\$P(z_t,q_t,\Delta t)$};
\node[sensor, above=of pred] (actions)
    {thruster sequence\\$u_{t:t+K-1}$};
\node[learned, right=7mm of pred] (heads)
    {future latent $\hat z_{t+K}$\\$\widehat{\Delta v}$, sonar profile};
\node[target, below=of heads] (ema)
    {EMA target encoder\\$z^+_{t+K}=\operatorname{sg}(\bar E(o_{t+K}))$};
\node[box, below=of pred, xshift=-20mm, fill=aquablue!4] (planner)
    {shared receding-horizon planner\\candidate cost $\rightarrow$ selected thruster command};

\draw[arrow] (cam) -- (ecam);
\draw[arrow] (sonar) -- (esonar);
\draw[arrow] (prop) -- (eprop);
\draw[arrow] (obs.east) -- (fusion.west);
\draw[arrow] (mask) -- (fusion);
\draw[arrow] (fusion) -- node[above] {$z_t$} (pred);
\draw[arrow] (actions) -- node[right] {GRU $E_u$} (pred);
\draw[arrow] (pred) -- (heads);
\draw[loss] (ema) -- node[right, text=aquagreen] {latent target} (heads);
\draw[arrow] (heads.south west) -- (planner.north east);
\draw[arrow] (planner.north west) -- (pred.south west);
\end{tikzpicture}
\caption{\method architecture and closed-loop use. Sensor-specific encoders
produce a mask-aware fused state. A GRU encodes future thruster commands, and
the predictor is trained against an EMA future target with physical auxiliary
heads. During closed-loop evaluation, the same predictor scores a shared
action library.}
\label{fig:architecture}
\end{figure*}
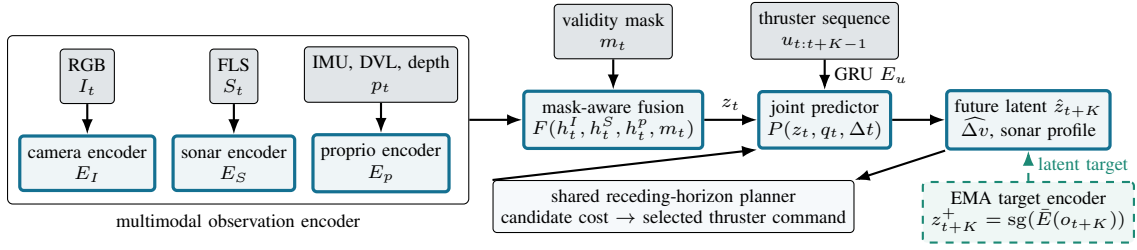

We propose \method, illustrated in Fig.~\ref{fig:architecture}. Sensor-specific
encoders feed a mask-aware fusion module. An action encoder summarizes a short
sequence of eight normalized thruster commands. The predictor learns against a
stop-gradient exponential-moving-average (EMA) target while physical heads
estimate future velocity change and near-range sonar intensity. A
counterfactual action margin encourages action sensitivity. \methodbase uses
complete sensor streams, \methodcamera retains camera and proprioception,
\methodsonar retains sonar and proprioception, and \methodrobust adds
camera/sonar modality dropout and DVL dropout. At test time, one evaluation
uses three-predictor ensembles; a seed-wise evaluation uses one
validation-selected checkpoint per seed.

This paper makes three contributions:

\begin{itemize}
    \item a sensor-configurable action-conditioned JEPA family with
    mask-aware fusion, physical auxiliary heads, and a control-sensitive
    latent objective, spanning full multimodal, camera-only, sonar-only, and
    sensor-dropout operating regimes;
    \item controlled comparisons with state-only prediction and established
    supervised and recurrent dynamics baseline families, implemented with the
    same data, action library, task cost, and planner, plus one-factor
    objective and corruption ablations; and
    \item a paired closed-loop evaluation across unseen maps, four visibility
    settings, dynamics shifts, and scheduled DVL loss, in which
    \methodbase achieves the strongest aggregate result and significantly
    improves both success and final error over state-only prediction, followed
    by a 2,160-run seed-wise analysis that isolates a supported endpoint-error
    advantage over the sonar-only family member.
\end{itemize}

The novelty is not a new sensor encoder, recurrent cell, or planner in
isolation. It is the sensor-configurable integration and controlled evaluation
of action-conditioned joint-embedding prediction for underwater robot dynamics
under structured partial observability.

The evaluation is organized around three questions. First, does the learned
future embedding distinguish consequences of different thruster sequences,
rather than merely extrapolate vehicle inertia? Second, which components of
the training objective preserve prediction when camera, sonar, or DVL
measurements disappear? Third, when the action library and task cost are held
fixed, does the representation improve goal reaching across unseen geometry,
optical attenuation, and dynamics shifts? We answer the first two with
held-out prediction and one-factor interventions, and the third with paired
closed-loop episodes.

\section{Related Work}

\subsection{Latent dynamics and model-based control}

Learned world models support planning by rolling compact states forward under
candidate actions. World Models separates visual compression, recurrent
dynamics, and control \cite{ha2018world}; PlaNet and Dreamer learn recurrent
latent state-space models for planning or imagined policy learning
\cite{hafner2019planet,hafner2020dreamer}. Probabilistic ensembles explicitly
represent dynamics uncertainty \cite{chua2018deep}, while MuZero and TD-MPC2
show that task-relevant latent prediction can avoid reconstructing the full
observation \cite{schrittwieser2020muzero,hansen2024tdmpc2}. These methods
establish strong action-conditioned alternatives to \method. Our recurrent and
supervised baselines test whether an underwater JEPA objective adds value when
the observation encoders, data, action horizon, and downstream planner are
held fixed.

\subsection{Joint-embedding predictive learning}

JEPA frames representation learning as prediction in an abstract feature
space \cite{lecun2022path}. I-JEPA predicts masked image-region embeddings
without negative pairs \cite{assran2023ijepa}, and V-JEPA extends the idea to
spatiotemporal video targets \cite{bardes2024vjepa}. These models motivate our
EMA target and feature-space loss, but visual temporal predictability alone
does not demonstrate controllability. \method conditions every rollout on the
proposed thruster sequence and adds a margin requiring the executed action to
match the future target better than inverse and zero-action alternatives.

\subsection{Multimodal learning under missing sensors}

Multimodal fusion must accommodate different geometries, sampling rates, and
failure modes \cite{baltrusaitis2019multimodal}. Modality dropout is a
well-established way to prevent over-reliance on a single stream
\cite{neverova2016moddrop}. Camera and FLS are especially complementary
underwater: the former carries appearance and the latter supplies acoustic
range structure when visibility deteriorates. Existing synchronized
sonar--visual collections, such as SOVIS \cite{chen2026sovis}, are valuable for
cross-modal perception, but SOVIS releases camera, FLS, temperature, and
pressure rather than action-labelled trajectories with IMU/DVL and causal
thruster commands. We therefore use simulator trajectories for action-conditioned
learning and treat real sonar--visual transfer as a separate open problem.

\subsection{Underwater simulation and planning}

Marine vehicles exhibit added mass, damping, buoyancy, currents, and actuator
allocation effects \cite{fossen2021handbook}. Gazebo-based UUV Simulator and
Stonefish provide physics and sensor interfaces for repeatable underwater
robot studies \cite{manhaes2016uuv,cieslak2019stonefish}. We use Stonefish
through the ROS/u0env environment released with USIM and U0
\cite{gu2025usimu0}, pinned at commit \texttt{3f4b840}. The base stack provides
camera, inertial and velocity measurements, pressure, thruster interfaces, and
privileged simulator state; our fail-closed collection extension additionally
records synchronized setpoints and Stonefish's native raw FLS stream. Our planner resembles
sampling-based model predictive control \cite{williams2018mppi}, but candidate
actions and costs are identical across learned models to isolate predictive
representations.

The closest conceptual alternatives therefore occupy different points on two
axes: what is predicted and how partial observations are handled. Recurrent
world models predict state-like features through sequential transitions;
supervised dynamics directly regress physical outcomes; and visual JEPAs
predict abstract targets without a control interface. Our comparison holds
the multimodal encoders, training trajectories, action horizon, and planner
interface as constant as possible. This makes the objective---joint-embedding,
supervised, or recurrent---the principal experimental difference rather than
the amount of privileged state or planner tuning.

\section{Problem Formulation}

Let $x_t$ denote the unobserved vehicle--environment state, including pose,
velocity, local geometry, current, and actuator effectiveness. The robot sees
only a multimodal observation $o_t$ and applies an eight-dimensional thruster
command $u_t$. Optical attenuation, acoustic artifacts, and intermittent DVL
lock make this a partially observed controlled process. We seek a compact
representation $z_t=E(o_t)$ whose predicted future $\hat z_{t+K}$ changes with
the proposed control sequence $u_{t:t+K-1}$ while retaining information needed
for short-horizon navigation.

This goal differs from reconstructing the next sensor frame. The future camera
contains illumination and texture changes that need not affect the best
thruster command, while the future sonar contains speckle and view-dependent
returns. We therefore train in representation space and attach only two
physical readouts used by the planner: velocity change and a coarse near-range
sonar profile. The target encoder is used during learning but is not required
for online action selection.

For a fixed candidate set $\mathcal U_t$, the downstream decision is
\begin{equation}
 u_t^*=\operatorname*{first}\arg\min_{\mathbf u\in\mathcal U_t}
 J\!\left(P(E(o_t),\mathbf u),o_t,g_t\right),
 \label{eq:decision}
\end{equation}
where $g_t$ is the navigation goal and only the first command is executed
before replanning. All learned baselines expose the same velocity and sonar
readouts to the same $J$. Consequently, closed-loop differences measure the
utility of their predictive representations within this shared control
interface, not superiority of a separately tuned controller.

\section{AquaJEPA}

\subsection{Observation and action model}

At time $t$, the robot receives
\begin{equation}
 o_t=(I_t,S_t,p_t,m_t),
\end{equation}
where $I_t$ is RGB, $S_t$ is raw FLS intensity,
$p_t\in\mathbb{R}^{15}$ contains DVL velocity, IMU angular velocity and linear
acceleration, IMU orientation, pressure, and DVL altitude, and
$m_t\in\{0,1\}^3$ marks camera, sonar, and proprioceptive availability.
Normalized thruster control $u_t\in[-1,1]^8$ is sampled at 10 Hz. The model
predicts $K$ steps ahead from $o_t$ and $u_{t:t+K-1}$.

\subsection{Mask-aware multimodal encoder}

Separate convolutional camera and sonar encoders and a two-layer
proprioceptive MLP produce 64-dimensional features
\begin{equation}
 h_t^I=E_I(I_t),\quad h_t^S=E_S(S_t),\quad h_t^p=E_p(p_t).
\end{equation}
Fusion explicitly receives validity:
\begin{equation}
 z_t=F([m_t^Ih_t^I;m_t^Sh_t^S;m_t^ph_t^p;m_t]).
 \label{eq:fusion}
\end{equation}
For \methodrobust only, camera or sonar is removed during training with
probability 0.35, and DVL velocity and altitude are independently dropped with
probability 0.35 and zeroed before proprioceptive encoding. \methodbase uses
the same masked fusion interface without training-time sensor dropout.

Both image encoders use four stride-two $5\times5$ convolutional blocks with
32, 64, 96, and 128 channels, GroupNorm, and GELU, followed by global average
pooling and a linear projection. RGB and FLS inputs are independently resized
to $96\times128$ and scaled to $[0,1]$. The proprioceptive encoder and fusion
network are two-layer MLPs with GELU and LayerNorm. The 15 proprioceptive
values comprise three DVL velocities, three IMU angular velocities, three IMU
linear accelerations, a four-dimensional orientation quaternion, pressure,
and DVL altitude. Multiplying each modality embedding by its mask before
concatenation prevents a missing sensor from being confused with an arbitrary
learned feature, while appending $m_t$ tells the fusion layer which evidence
was actually available.

\subsection{Action-conditioned joint prediction}

A GRU summarizes the $K=5$ thruster commands as $q_t$. The predictor consumes
the current embedding, action summary, and physical horizon:
\begin{equation}
 \hat z_{t+K}=P([z_t;q_t;\Delta t]).
 \label{eq:predictor}
\end{equation}
An EMA copy of the multimodal encoder builds a stop-gradient target
$z^+_{t+K}=\operatorname{sg}(\bar E(o_{t+K}))$ with momentum 0.99. Auxiliary
heads decode future velocity change $\widehat{\Delta v}_{t+K}$ and a 32-bin
sonar range-intensity profile $\hat s_{t+K}$.

The loss is
\begin{equation}
\begin{split}
 \mathcal L={}&\mathcal L_{\rm lat}+\mathcal L_{\rm vel}
 +0.5\mathcal L_{\rm sonar}+0.1\mathcal L_{\rm cm}
 +2.0\mathcal L_{\rm act},
 \label{eq:loss}
\end{split}
\end{equation}
where latent and cross-modal terms use cosine distance and the physical heads
use $L_1$ losses. For latent prediction error
$e(u)=1-\cos(P(z_t,u,\Delta t),z^+_{t+K})$, the action loss is
\begin{equation}
 \mathcal L_{\rm act}=\tfrac12\sum_{\tilde u\in\{-u,0\}}
 [0.02+e(u)-e(\tilde u)]_+.
 \label{eq:margin}
\end{equation}
This explicitly penalizes locally action-invariant predictions.

The cross-modal term aligns the camera and sonar embeddings at the current
time, whereas the latent term constrains the fused future. These terms play
different roles: cross-modal alignment supplies a common geometric signal
when one exteroceptive stream is weak, while the future target preserves
information not captured by the low-dimensional physical heads. The action
margin is evaluated only for samples with non-negligible commands and compares
the executed sequence with both its inverse and a zero sequence. Thus a model
cannot satisfy the objective solely by copying $z_t$ or predicting average
inertial drift.

\subsection{Shared receding-horizon planner}

At 2 Hz, every learned method scores the same goal-directed, constant, pulse,
ramp, and braking action candidates over a 0.5 s horizon. A three-seed ensemble
minimizes
\begin{equation}
\begin{split}
 J(u)={}&\lambda_v\|\hat v_{t+K}(u)-v_t^*\|_1
 +\lambda_c C_{\rm near}(u)+\lambda_e\|u\|_1\\
 &+\lambda_s\|u_t-u_{t-1}\|_1
 +\lambda_q\operatorname{Std}[\hat v_{t+K}(u)].
\end{split}
\end{equation}
The desired body velocity $v_t^*$ is computed from the navigation goal.
Action limits, task cost, online residual correction, and success radius are
shared across methods.

The candidate library contains neutral and previous-action holds, a braking
pulse, axis-aligned surge/sway/yaw/heave primitives, and goal-aligned mixtures with
hold, pulse, and ramp temporal profiles. Primitive amplitudes use scales
$\{0.60,0.85,1.00,1.15\}$ around 0.30 and are clipped at 0.45. The weights are
$\lambda_v=1$, $\lambda_c=0.05$, $\lambda_e=0.001$,
$\lambda_s=0.005$, and $\lambda_q=0.05$. The sonar cost averages the nearest
third of the predicted 32-bin profile; uncertainty is the ensemble standard
deviation of predicted final velocity. When DVL is available, an exponential
residual estimate ($\alpha=0.25$, clipped to $0.20$ m/s) corrects systematic
velocity bias for every learned method.

At each planner update, the goal displacement is rotated into the DVL body
frame and capped at 0.40 m/s. Every ensemble member rolls out every candidate,
the costs are averaged, and the first thruster vector of the minimum-cost
sequence is held until the next 2 Hz update. During a declared DVL-loss
window, velocity and altitude inputs are zeroed, the planner's velocity-valid
flag is updated,
and residual adaptation is paused. This causal loop uses no privileged
odometry inside the predictive model; simulator odometry is used only to form
the navigation target and compute evaluation metrics.

\section{Experimental Evaluation}

\subsection{Training data and implementation details}

We collect 150 synchronized 30 s Stonefish BlueROV2 episodes at 10 Hz: 120
training episodes (60 min), 15 validation episodes (7.5 min), and 15 held-out
test episodes (7.5 min). Complete environment episodes, rather than frames,
define splits. The collector requires camera, raw FLS, IMU, DVL, pressure, and
the applied eight-thruster command, and rejects non-monotonic or
out-of-schedule samples. Randomization covers four obstacle-layout families,
initial pose, lighting, four water-visibility coefficients, current magnitude
and direction, obstacle jitter, and action amplitude. The 0.18 m/s current and
actuator-efficiency shift used in closed loop is excluded from training.

Training examples are constructed as causal five-command windows: the current
observation precedes the recorded command sequence and the target observation
is taken after the fifth command. The physical elapsed time is included in the
predictor rather than assumed from an array index. Sonar supervision is formed
by averaging intensity over bearing and adaptively pooling range to 32 bins;
velocity supervision is the change from the current to the future DVL reading.
No frame from a validation or test episode appears in training.

All models use the same image sizes ($96\times128$), 64-dimensional latent,
18 AdamW epochs, batch size 64, learning rate $3\times10^{-4}$, weight decay
$10^{-4}$, mixed precision, and seeds 11, 22, and 33. The nine methods and
ablations yield 27 independent training runs. Checkpoints are selected using
validation data only, and the selected models are carried unchanged into
closed-loop evaluation.

Validation checkpoint selection is also objective-aware. JEPA variants use a
composite of future-velocity MAE, executed-versus-zero-action ranking, and the
signed no-action gap; rankings below 0.55 and negative gaps incur penalties.
Supervised and recurrent baselines are selected by future-velocity MAE. The
held-out test split is used only for final reporting. This objective-aware
selection avoids choosing a low-error but action-insensitive checkpoint for
\method while retaining the natural supervised criterion for the physical
dynamics baselines.

\subsection{Family configurations, baselines, and ablations}

The proposed family contains four sensor configurations. \methodbase uses
camera, sonar, and proprioception without training-time sensor dropout.
\methodcamera permanently masks sonar, while \methodsonar permanently masks
camera; both retain proprioception and the same action-conditioned JEPA
objective. \methodrobust retains full multimodal input but adds 0.35
camera/sonar modality dropout and 0.35 DVL dropout during training. All members
share the encoder dimensions, action GRU, EMA target, physical heads, action
margin, validation-only checkpoint rule, and downstream planner. The ensemble
evaluation compares \methodbase and \methodrobust; the seed-wise evaluation adds
\methodcamera and \methodsonar to isolate sensor configuration.

The closed-loop controls are: (i) a reactive controller without a learned
model; (ii) a state-only predictor that retains proprioception but masks both
exteroceptive streams; (iii) supervised action-conditioned dynamics, which
shares the encoders and predictor but optimizes only velocity and sonar-profile
targets; and (iv) a recurrent supervised world model that applies a GRUCell
transition for each action. The latter two are controlled implementations of
established non-JEPA baseline families, not externally sourced checkpoints.
All learned methods use the same downstream planner. The ensemble comparison
matches \methodrobust's sensor-corruption schedule; the seed-wise study also
trains stronger zero-dropout supervised and recurrent controls.

Four one-factor ablations replace the EMA target by an online stop-gradient
target, remove the action margin, hide the explicit mask while retaining
zeroed inputs, or remove camera/sonar modality dropout while retaining DVL
dropout. Ablations use the same three model seeds and held-out prediction
episodes and are evaluated with the shared planner.

We evaluate sensor-loss robustness for \methodrobust and the no-explicit-mask
and no-modality-dropout ablations on the held-out test episodes under nominal,
camera-absent, sonar-absent, DVL-absent, and combined camera--DVL conditions.
We report five-step future-velocity MAE across the three model seeds.

\subsection{Factorial closed-loop evaluation}

The closed-loop benchmark contains 120 paired scenarios: five independent
environment-seed replicates of the 24-cell Cartesian product of three unseen
obstacle layouts (open arc, east corridor, and north slalom), four Stonefish
Jerlov coefficients (0.10, 0.15, 0.20, and 0.25), and nominal or shifted
dynamics. The shifted regime raises current magnitude from 0.08 to 0.18 m/s
and sets the four horizontal-thruster efficiencies to 0.82--0.88. Initial
pose, current direction, lighting, map, and dynamics are identical across
methods within an episode.

Each 55 s run removes DVL velocity and altitude during 10--15, 25--30, and
40--45 s. A 1 m terminal goal radius defines success. Every method is evaluated
on the same scenarios, producing \TotalClosedLoopRuns{} runs and 120 paired
inferential units.

To separate family behavior from ensemble averaging, the seed-wise evaluation
evaluates one validation-selected checkpoint for each of seeds 11, 22, and 33
on the same 120-scenario factorial design. It compares \methodbase,
\methodcamera, and \methodsonar with state-only and matched zero-dropout
supervised and recurrent controls. Pairing is preserved within each training
seed, yielding \SeedwiseClosedLoopRuns{} closed-loop runs.

\begin{table}[t]
\caption{Closed-loop evaluation design. Every method receives the same factor
cells, action candidates, task cost, and DVL-loss schedule.}
\label{tab:evaluation-design}
\centering
\small
\begin{tabular}{ll}
\toprule
Item & Setting \\
\midrule
Simulator / vehicle & Stonefish / BlueROV2, 10 Hz \\
Planner & 2 Hz; 5 steps / 0.5 s \\
Episode / success & 55 s / terminal error $\leq1$ m \\
Layouts & open arc, east corridor, north slalom \\
Visibility coefficient & 0.10, 0.15, 0.20, 0.25 \\
Current & 0.08 m/s nominal; 0.18 m/s shifted \\
Shifted actuation & horizontal efficiencies 0.82--0.88 \\
DVL missing & 10--15, 25--30, 40--45 s \\
Inference units & $5\times(3\times4\times2)=120$ paired episodes \\
\bottomrule
\end{tabular}
\end{table}

\subsection{Metrics and uncertainty}

Prediction outcomes are future-velocity MAE, sonar-profile MAE, latent cosine
error, and executed-versus-zero-action ranking. Closed-loop outcomes are
success, final and mean goal error, collision, minimum clearance, and command
total variation. We report every paired final-error difference and summarize
\methodbase minus each comparator by its mean, win/tie/loss count, and
percentile 95\% interval from 10,000 cell-stratified paired bootstrap
resamples. Results are also stratified by layout, visibility coefficient, and
dynamics regime.

Each bootstrap draw resamples five entire paired episodes within each of the
24 factor cells and then averages the cell means with equal weight; it never
resamples frames or control steps.
For final error, a negative difference means that \methodbase finishes closer to
the goal. A win is an episode with lower final error than the comparator; for
success, the reported difference is \methodbase's paired success indicator
minus the comparator indicator. These episode-level estimands match the factor sweep
and avoid treating highly correlated 10 Hz observations as independent.

For the seed-wise study, a 10,000-draw hierarchical paired bootstrap first
resamples training seeds and then resamples five paired environment replicates
within each factor cell. Its inferential unit is therefore a paired environment
episode nested within training seed, rather than a controller step or frame.

State-only isolates whether exteroceptive input adds value beyond
proprioception. We report success and final error as separate paired outcomes;
intervals excluding zero indicate supported paired differences. Collision and
clearance retain their safety-diagnostic role, and all other pairwise
comparisons are reported in full.

\section{Results}

\subsection{Hour-scale predictive evaluation}

The 15-episode held-out split is environment-disjoint from both training and
validation. It evaluates whether the expanded corpus preserves counterfactual
action sensitivity and missing-input robustness independently of the
closed-loop benchmark.

\subsection{Supervised baselines and component ablations}

Table~\ref{tab:prediction-ablation} compares both \method configurations,
dynamics baselines, and single-component ablations at the five-step horizon.
\PredictionResultSentence{}
For supervised models, latent action rank is a diagnostic against the stop-gradient
encoder target rather than a training objective.

\begin{table}[t]
\caption{Five-step held-out prediction (mean $\pm$ standard deviation across
three seeds). Rank means the executed action has lower target error than zero
action; ``w/o'' rows ablate \methodrobust.}
\label{tab:prediction-ablation}
\centering
\small
\resizebox{\columnwidth}{!}{\begin{tabular}{lccc}
\toprule
Method & Velocity MAE $\downarrow$ & Sonar MAE $\downarrow$ & Action rank $\uparrow$ \\
\midrule
State-only & 0.0029$\pm$0.0001 & 0.0107$\pm$0.0004 & 0.694$\pm$0.064 \\
AquaJEPA-base & 0.0030$\pm$0.0001 & 0.0062$\pm$0.0011 & 0.647$\pm$0.062 \\
Supervised dynamics & 0.0023$\pm$0.0000 & 0.0038$\pm$0.0001 & 0.374$\pm$0.073 \\
Recurrent world model & 0.0024$\pm$0.0001 & 0.0039$\pm$0.0003 & 0.208$\pm$0.028 \\
AquaJEPA-robust & 0.0032$\pm$0.0002 & 0.0070$\pm$0.0018 & 0.634$\pm$0.052 \\
w/o EMA target & 0.0031$\pm$0.0003 & 0.0105$\pm$0.0008 & 0.682$\pm$0.072 \\
w/o action margin & 0.0030$\pm$0.0001 & 0.0057$\pm$0.0002 & 0.573$\pm$0.033 \\
w/o explicit masks & 0.0032$\pm$0.0001 & 0.0063$\pm$0.0005 & 0.675$\pm$0.051 \\
w/o modality dropout & 0.0032$\pm$0.0001 & 0.0063$\pm$0.0004 & 0.599$\pm$0.057 \\
\bottomrule
\end{tabular}
}
\end{table}

\CorruptionResultSentence

\begin{table}[t]
\caption{Future-velocity MAE under controlled sensor interventions (mean
$\pm$ standard deviation over three model seeds). DVL dropout remains enabled
during training for all rows.}
\label{tab:corruption-ablation}
\centering
\small
\resizebox{\columnwidth}{!}{\begin{tabular}{lccc}
\toprule
Method & Nominal & Camera absent & Camera+DVL absent \\
\midrule
AquaJEPA-robust & 0.0032$\pm$0.0002 & 0.0032$\pm$0.0002 & 0.0037$\pm$0.0001 \\
robust w/o masks & 0.0032$\pm$0.0001 & 0.0032$\pm$0.0001 & 0.0037$\pm$0.0001 \\
robust w/o modality dropout & 0.0032$\pm$0.0001 & 0.0088$\pm$0.0023 & 0.0090$\pm$0.0023 \\
\bottomrule
\end{tabular}
}
\end{table}

\subsection{Closed-loop evaluation under visibility and dynamics shifts}

\begin{figure*}[!b]
\centering
\includegraphics[width=0.88\textwidth]{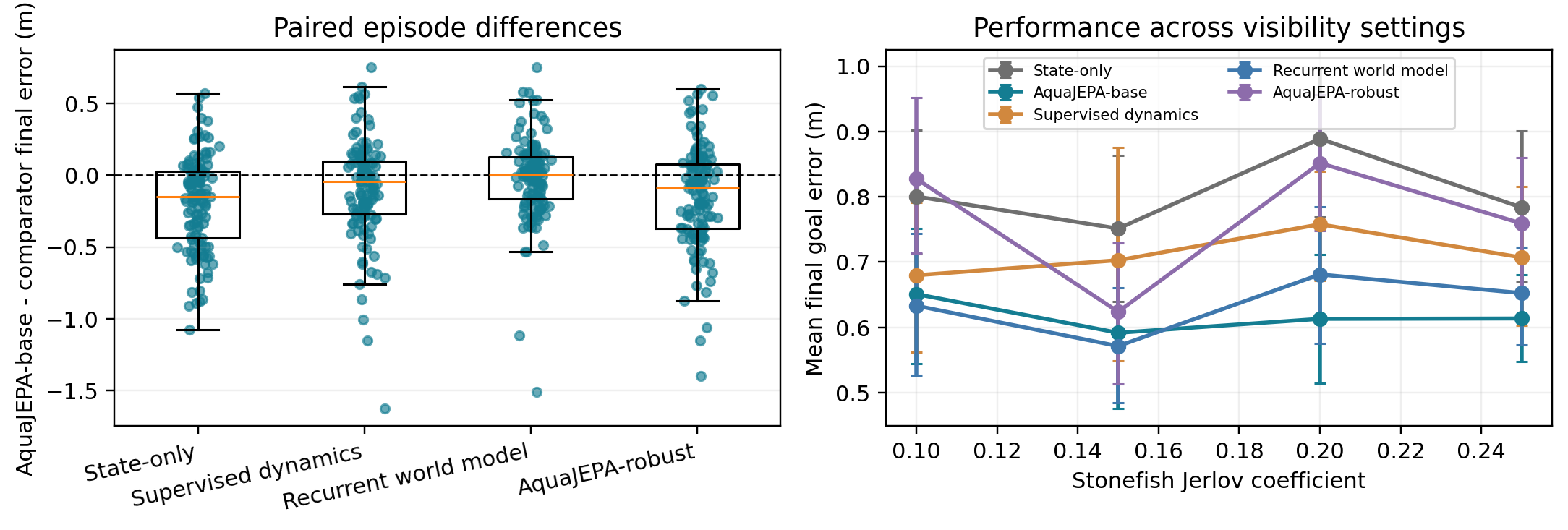}
\caption{Closed-loop behavior under partial observability. Left: per-episode
paired final-error differences between \methodbase and each learned comparator;
values below zero favor \methodbase, boxes show interquartile ranges, and dots
show all 120 episodes. Right: mean final error for both \method configurations
and the learned baselines at each Stonefish visibility coefficient, pooling
three layouts and two dynamics regimes; error bars are paired-cell bootstrap
95\% intervals.}
\label{fig:paired-errors}
\end{figure*}

Table~\ref{tab:closed-loop} reports all six planners on the 120-episode
matrix. \ClosedLoopResultSentence{} The paired distribution in
Fig.~\ref{fig:paired-errors} shows whether an aggregate difference is driven by
a few environments or is repeated across the factor sweep.

\begin{table}[t]
\caption{Closed-loop results across 120 paired environments. Final error and
clearance are means in meters; collision is the percentage of episodes.}
\label{tab:closed-loop}
\centering
\small
\resizebox{\columnwidth}{!}{\begin{tabular}{lrrrr}
\toprule
Method & Success & Collision & Final error $\downarrow$ & Clearance $\uparrow$ \\
\midrule
Reactive & 31/120 & 0.0\% & 1.374 & 1.223 \\
State-only & 76/120 & 0.0\% & 0.806 & 0.629 \\
AquaJEPA-base & 91/120 & 0.0\% & 0.617 & 0.542 \\
Supervised dynamics & 87/120 & 0.0\% & 0.712 & 0.479 \\
Recurrent world model & 87/120 & 0.0\% & 0.634 & 0.491 \\
AquaJEPA-robust & 84/120 & 0.0\% & 0.766 & 0.560 \\
\bottomrule
\end{tabular}
}
\end{table}

Table~\ref{tab:paired-effects} reports the episode-level estimands behind the
aggregate means. Against state-only, \methodbase wins 84 of 120 paired
final-error comparisons and both its final-error and success intervals exclude
zero. Its final-error interval also favors it over supervised dynamics, while
the success interval includes zero. Both outcomes are unresolved against the
recurrent world model. Relative to \methodrobust, \methodbase has lower final
error and higher success, exposing the aggregate-performance side of the
sensor-dropout trade-off.

\begin{table*}[t]
\caption{Paired closed-loop effects for \methodbase minus each comparator
across 120 episodes. Final-error differences are in meters; negative is
better. W/T/L counts lower/equal/higher final error for \methodbase. Success
differences are paired proportions. Intervals use 10,000 cell-stratified
bootstrap resamples.}
\label{tab:paired-effects}
\centering
\small
\resizebox{0.93\textwidth}{!}{\begin{tabular}{lccc}
\toprule
Comparator & $\Delta$ final error [95\% CI] & W/T/L & $\Delta$ success [95\% CI] \\
\midrule
Reactive & $-0.756$ [$-0.802,-0.710$] & 117/0/3 & $+0.500$ [$+0.442,+0.558$] \\
State-only & $-0.189$ [$-0.239,-0.140$] & 84/0/36 & $+0.125$ [$+0.075,+0.183$] \\
Supervised dyn. & $-0.095$ [$-0.150,-0.041$] & 69/0/51 & $+0.033$ [$-0.017,+0.083$] \\
Recurrent WM & $-0.017$ [$-0.063,+0.028$] & 60/0/60 & $+0.033$ [$-0.008,+0.083$] \\
AquaJEPA-robust & $-0.148$ [$-0.202,-0.095$] & 77/0/43 & $+0.058$ [$+0.017,+0.108$] \\
\bottomrule
\end{tabular}
}
\end{table*}

\subsection{Where the closed-loop difference appears}

\StratifiedResultParagraph

The visibility sweep is deliberately not a clean-camera versus turbid-camera
comparison: every cell is already a degraded underwater setting and every
episode contains 15 s of scheduled DVL loss. The right panel of
Fig.~\ref{fig:paired-errors} tests whether the state-only success difference
persists as optical attenuation changes while sonar remains available. It does:
\methodbase exceeds state-only at all four coefficients and attains the lowest
mean final error at three of four settings while remaining near-best at 0.10.

Dynamics shift is the stronger disturbance: \methodbase's mean final error
rises from 0.261 m in nominal cells to 0.974 m with increased current and
reduced horizontal-thruster effectiveness. It succeeds in 60/60 nominal cells
and retains 31/60 successes after the shift, compared with 58/60 and 18/60 for
state-only, 59/60 and 25/60 for \methodrobust, and 59/60 and 28/60 for both
supervised dynamics and the recurrent model.
No planner collides in the benchmark, so collision rate does not explain
the success difference and is not claimed as a comparative advantage; complete
clearance results remain in Table~\ref{tab:closed-loop}.

\subsection{Seed-wise AquaJEPA family analysis}

Table~\ref{tab:seedwise-family} reports the separate single-checkpoint study.
Across 360 episodes, \methodbase reaches 219 goals (60.8\%), finishes at
0.865~m mean final error, and records no collisions. Relative to
\methodsonar, it reduces paired final error by 0.118~m (hierarchical paired
95\% CI: 0.047--0.181~m) and wins 219 of 360 endpoint-error comparisons. The
reduction is repeated for every training seed: 0.178, 0.077, and 0.099~m for
seeds 11, 22, and 33. \methodcamera remains a strong camera-first member of
the family; this analysis therefore supports the acoustic-only comparison
rather than a universal fusion-superiority claim.

\begin{table}[t]
\caption{Single-checkpoint robustness across three training seeds (360 paired
episodes per controller). Success and collision are counts; final error is the
mean in meters. Bold marks the best AquaJEPA-family aggregate.}
\label{tab:seedwise-family}
\centering
\small
\resizebox{\columnwidth}{!}{\begin{tabular}{lrrr}
\toprule
Method & Success & Final error $\downarrow$ & Collision \\
\midrule
\multicolumn{4}{l}{\emph{AquaJEPA family}} \\
AquaJEPA-base & \textbf{219/360} & \textbf{0.865} & 0/360 \\
AquaJEPA-C (camera) & 197/360 & 0.948 & 1/360 \\
AquaJEPA-S (sonar) & 202/360 & 0.983 & 1/360 \\
\addlinespace
\multicolumn{4}{l}{\emph{Controls and non-JEPA baselines}} \\
State-only & 196/360 & 0.993 & 0/360 \\
Supervised dynamics & 238/360 & 0.712 & 0/360 \\
Recurrent world model & 257/360 & 0.710 & 0/360 \\
\bottomrule
\end{tabular}
}
\end{table}

\section{Discussion}

The study separates three explanations that a single multimodal result cannot
resolve. The state-only baseline tests whether direct motion
sensing is sufficient; supervised dynamics tests whether physical auxiliary
targets alone explain planning quality; and the recurrent model tests whether
an autoregressive transition is enough. The one-factor ablations then locate
which parts of the JEPA training recipe contribute to prediction under sensor
loss.

The experiments also show why a supervised prediction table alone is
insufficient for model selection. \method's action margin strengthens ranking
of the executed action against matched counterfactuals, which directly serves
the shared planner's comparison of unexecuted candidates. Coupling this
control-sensitive latent with physical velocity and sonar heads makes the
objectives complementary: one separates candidate futures, while the other
preserves calibrated short-horizon predictions.

The missing-sensor intervention gives a similarly specific mechanism result.
Relative to its matched no-modality-dropout ablation, \methodrobust more than
halves velocity error when camera is absent, including the combined
camera--DVL condition. Across the controlled corruptions, training exposure to
missing modalities is therefore the clearest isolated driver of test-time
sensor-loss robustness.

The paired factorial design holds the map, water coefficient, current,
actuator efficiencies, initial pose, planner, and DVL-loss schedule fixed
within each episode. The hour-scale study therefore makes exteroception
behaviorally relevant under controlled partial observability: the success
and final-error intervals for \methodbase against state-only exclude zero, and
the advantage persists across all visibility levels. The per-episode analysis
also exposes heterogeneity hidden by aggregate means and localizes the gain to
the learned predictive representation rather than easier random draws or a
separately tuned controller.

The complete baseline matrix shows that the strongest aggregate controller is
the base configuration of the proposed AquaJEPA family, not an independent
multimodal baseline. \methodbase significantly reduces final error relative to
state-only and supervised dynamics, while its comparison with the recurrent
model is unresolved. The base--robust contrast then isolates a practical
trade-off: explicit sensor-loss training improves blackout robustness but
reduces pooled closed-loop performance in this evaluation.

The family view adds a deployment interpretation to these comparisons.
\methodbase is the full-sensor member, \methodcamera and \methodsonar retain
the same action-conditioned objective for sensor-constrained platforms, and
\methodrobust targets anticipated blackouts. The seed-wise reduction against
\methodsonar shows that joint optical--acoustic input improves endpoint
accuracy over acoustic-only prediction without depending on a particular
training seed.

\section{Limitations}
\label{sec:limitations}

Three limitations bound the claim. First, all evidence comes from one Stonefish
BlueROV2 model with a discrete action library, privileged scoring odometry, and
approximate collision geometry. Second, the three-seed single-checkpoint study
supports a sensor-specific comparison rather than universal family or objective
superiority; its matched supervised and recurrent controls achieve higher
aggregate success. Third,
Jerlov coefficients and injected DVL/actuator failures are controlled proxies
rather than a complete model of real water, acoustic artifacts, timing drift,
or hardware degradation. Real action-conditioned validation requires
synchronized camera, sonar, state, and causal thruster telemetry.

\section{Conclusion}

\method provides a sensor-configurable family of future representations
conditioned on thruster sequences and uses their physical heads for
receding-horizon action selection. The expanded simulation study compares
strong state-only, supervised, and recurrent alternatives, isolates four
design components, and tests fresh paired episodes across maps, visibility,
dynamics, and DVL loss. \ConclusionResultSentence{} The next step is
evaluation on real trajectories with verified command timing.

\section*{Acknowledgment}

OpenAI Codex was used under author supervision for
language editing throughout all sections of the manuscript.
The authors verified all claims, citations, analyses, code, and
numerical results.

\bibliographystyle{IEEEtran}
\balance
\bibliography{references,references_extra}

\end{document}